\documentclass[journal]{IEEEtranTIE}
\usepackage{graphicx}
\usepackage{cite}
\usepackage{picinpar}
\usepackage{amsfonts,amssymb,amsthm,amsmath}
\usepackage{url}
\usepackage{flushend}
\usepackage[latin1]{inputenc}
\usepackage{colortbl}
\usepackage{soul}
\usepackage{multirow}
\usepackage{pifont}
\usepackage{color}
\usepackage{alltt}
\usepackage[hidelinks]{hyperref}
\usepackage{enumerate}
\usepackage{siunitx}
\usepackage{breakurl}
\usepackage{epstopdf}
\usepackage{pbox}

\newtheorem{theorem}{Theorem}
\newtheorem{lemma}{Lemma}
\newtheorem{remark}{Remark}
\newtheorem{assumption}{Assumption}
\newtheorem{definition}{Definition}

\def\bs{\boldsymbol s}
\def\bu{\boldsymbol u}
\def\bx{\boldsymbol x}
\def\bp{\boldsymbol p}
\def\bh{\boldsymbol h}
\def\bs{\boldsymbol s}
\def\be{\boldsymbol e}

\def\bw{\boldsymbol w}
\def\bp{\boldsymbol p}
\def\b\alpha{\boldsymbol \alpha}

\begin{document}
\title{	Multiple Non-cooperative Targets Encirclement by Relative Distance based Positioning and Neural Anti-Synchronization Control}

\author{
	\vskip 1em
	
	Fen Liu, Shenghai Yuan, Wei Meng, Rong Su, \emph{Senior Member, IEEE}, Lihua Xie, \emph{{Fellow, IEEE}}

	\thanks{
		This work was partially supported by the National Natural Science Foundation of China (U21A20476, 62033003), Guangdong Introducing Innovative and Entrepreneurial Teams (2019ZT08X340) of Guangdong Province, the Local Innovative and Research Teams Project of Guangdong Special Support Program (2019BT02X353), the Agency for Science, Technology and Research (A*STAR) under its IAF-ICP Programme I2001E0067, via the COSMO WP5 project, the National Research Foundation, Singapore, under its Medium-Sized Center for Advanced Robotics Technology Innovation (CARTIN).
		
		F. Liu, W. Meng are with Guangdong Provincial Key Laboratory of Intelligent Decision and Cooperative Control, School of Automation, Guangdong University of Technology, Guangzhou 510006, (e-mail: 1112004018@mail2.gdut.edu.cn, meng0025@ntu.edu.sg).
		
		S. Yuan, R. Su, L. Xie are with the School of Electrical and Electronic Engineering, Nanyang Technological University, Singapore 639798, Singapore (e-mail:shyuan@ntu.edu.sg, rsu@ntu.edu.sg, elhxie@ntu.edu.sg).
	}
}
\maketitle

\begin{abstract}
From prehistoric encirclement for hunting to GPS orbiting the earth for positioning, target encirclement has numerous real world applications. However, encircling multiple non-cooperative targets in GPS-denied environments remains challenging. In this work, multiple targets encirclement by using a minimum of two tasking agents, is considered where the relative distance measurements between the agents and the targets can be obtained by using onboard sensors. Based on the measurements, the center of all the targets is estimated directly by a fuzzy wavelet neural network (FWNN) and the least squares fit method. Then, a new distributed anti-synchronization controller (DASC) is designed so that the two tasking agents are able to encircle all targets while staying opposite to each other. In particular, the radius of the desired encirclement trajectory can be dynamically determined to avoid potential collisions between the two agents and all targets. Based on the Lyapunov stability analysis method, the convergence proofs of the neural network prediction error, the target-center position estimation error, and the controller error are addressed respectively. Finally, both numerical simulations and UAV flight experiments are conducted to demonstrate the validity of the encirclement algorithms. The flight tests recorded video and other simulation results can be found in \textcolor[rgb]{0.00,0.0,1.00}{https://youtu.be/B8uTorBNrl4}.
\end{abstract}


\markboth{IEEE TRANSACTIONS ON INDUSTRIAL ELECTRONICS}%
{}

\definecolor{limegreen}{rgb}{0.2, 0.8, 0.2}
\definecolor{forestgreen}{rgb}{0.13, 0.55, 0.13}
\definecolor{greenhtml}{rgb}{0.0, 0.5, 0.0}

\vspace{-0.2cm}
\section{Introduction}

\IEEEPARstart{I}{n} recent years, target tracking and monitoring algorithms have been widely used in maritime, aerospace, environmental surveys, convoy escorts, and security patrol scenarios \cite{hung2022cooperative, ye2022multi,lv2022event}. The encirclement control, as a type of tracking and monitoring strategy, requires the mobile tasking agents to be able to circumnavigate the target while tracking \cite{zhang2022multi, peng2020event,yu2019cooperative}. Recently, many methods have been proposed along on this research direction \cite{liu2023moving,jiang2020line,yu2018cooperative,peng2020moving,dou2021moving}.
For example, in \cite{jiang2020line}, an estimator-based enclosing controller was proposed to drive an autonomous surface vehicle to surround and monitor a single target affected by random ocean currents. In addition to a single agent encircling a single target, multiple agents have also been used to encircle a single target by designing a corresponding consensus algorithm. In \cite{yu2018cooperative}, a dynamic control law was proposed that can enable multiple vehicles to move around a target along an ideal trajectory while the target is moving at a time-varying speed. In \cite{dou2021moving}, a relative position-based controller was proposed to enable a group of moving agents to surround a single moving target.

Compared with single target encirclement, the multiple target encirclement problem is much more complex. In recent works, there are two main ways to realize the multiple target encirclement. One is to let the tasking agents encircle each target separately, that is, to transform the problem of multiple targets encirclement into a single target encirclement \cite{zhang2022multi}. The other is to make all tasking agents surround all the targets with a desired trajectory \cite{dong2017necessary}. The desired trajectory of the encirclement task needs to be further designed carefully such that the agents do not collide with any of the targets\cite{matveev2017range,deghat2015multi}. In \cite{matveev2017range}, the radius of the trajectory was adjusted directly according to the measured distances between the agents and targets. In \cite{deghat2015multi}, the position of each target was obtained separately, and then the center and radius of trajectory needed by the agent to circumnavigate all targets are designed according to the obtained position.

Most of the mentioned target encirclement algorithms are based on the assumption that the targets are cooperative. For cooperative targets, the agents can directly obtain the feedback information from the target, such as position, displacement, etc \cite{Zuo2019Time,ju2021enclosing}. However, in real cases, the targets are non-cooperative, that is, the real-time states and motion models of the targets are all unknown. In some existing results, it is assumed that the non-cooperative target is stationary, or moving at a low or constant speed, thus ensuring that the model of the target can be obtained \cite{shames2011circumnavigation,hashemi2015unmanned,kou2022cooperative}. On the basis of this assumption, the distance and angle based localization algorithms have been proposed to solve the positioning problem of non-cooperative targets \cite{Thien2020Persistently,dong2020target}. For example, in \cite{dong2020target}, a mobile robot was assumed to have a known linear velocity, and a backstepping-based controller was designed to control the direction of the robot so as to achieve the encirclement of the unknown stationary target. However, for applications in complex environments, the non-cooperative target may be moving freely. In this case, the positioning of the targets still remains challenging. In addition, for the encirclement of multiple non-cooperative targets, positioning each target separately may lead to longer execution time and inaccuracy of the whole encirclement algorithm due to the accumulation of errors.

Generally, in the encirclement task, a single tasking agent can only observe parts of the states of the targets and cannot achieve the necessary coverage to the targets. Hence, for the monitoring task, multiple tasking agents are required to encircle and monitor the targets. However, other problems may occur such as the complexity of control algorithm, collision avoidance issues, and the implementation cost. To solve above issues, in this paper, we are dedicated to making a trade off, that is to adopt a minimum of two tasking agents to encircle freely moving multiple targets while achieving maximum sensor coverage to the targets.

\begin{figure}
\centering
  \includegraphics[width=6cm]{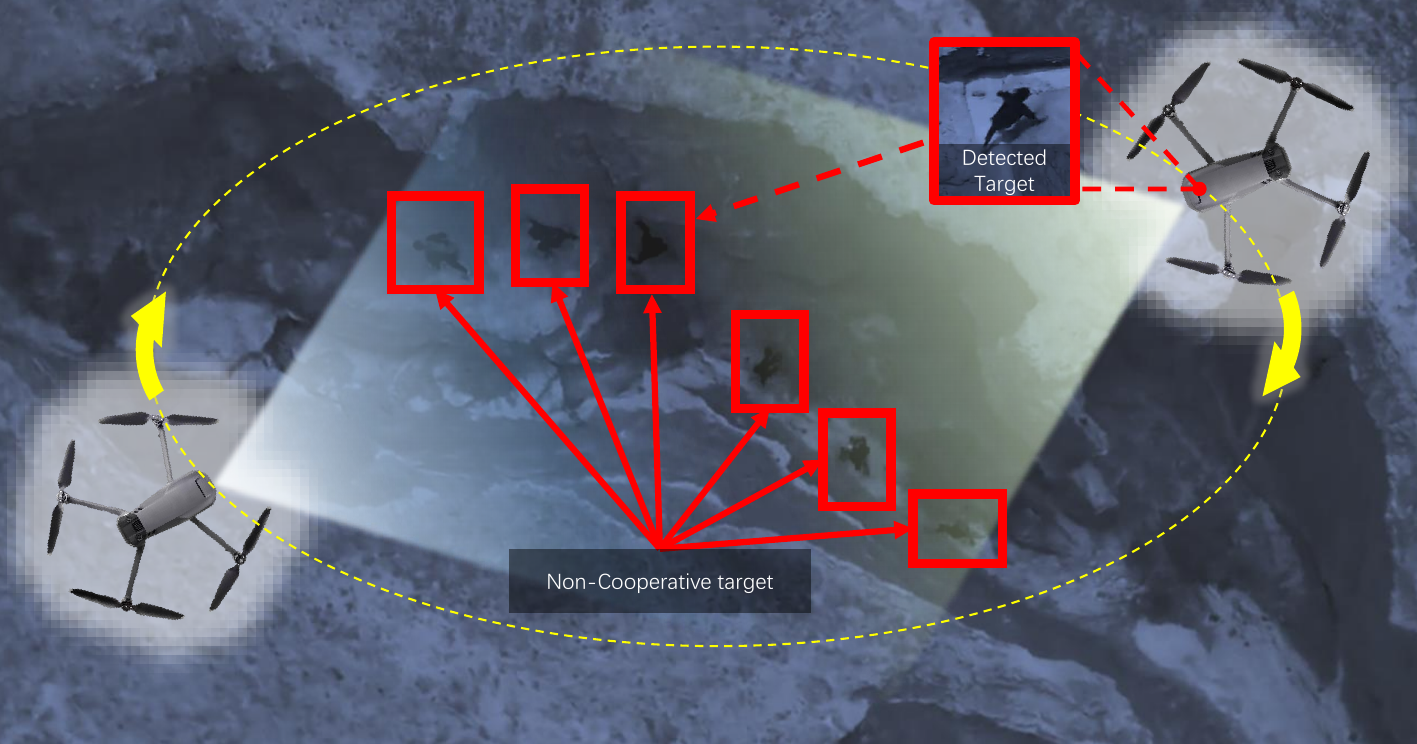}
 \caption{Multiple unknown and non-cooperative targets encircling by using only two tasking agents.}
  \label{AS}
\end{figure}
\setlength{\belowcaptionskip}{-0.5cm}

The research highlight of our work is shown in Fig. \ref{AS}, where multiple targets are successfully encircled by using only two mobile tasking agents. Moreover, we also assume that the GPS signal is not available in the tasking environments.  All targets we consider are completely unknown and non-cooperative, that is, the two agents cannot directly obtain the states of all targets. Given the challenges in practical applications, such as energy constraints, it is assumed that each agent can only carry a distance measurement sensor, such as vision sensor, attenna array, etc. Based on the relative distance information measurements among the tasking agents and the targets, an estimator is designed such that the target-center position can be estimated in real-time. Furthermore, based on the estimated position information, a distributed anti-synchronization controller (DASC) is designed so that the two tasking agents can symmetrically circumnavigate all the targets by dynamically adjusting the radius of the movement trajectories. The main contributions of this work are highlighted as follows.
\begin{enumerate}
 \item Different from the existing work related to targets encirclement \cite{yu2019cooperative,peng2021Cooperative}, the main focus in this work is to study the non-cooperative targets encirclement problem, that is, all state information about the targets cannot be obtained directly and the model of target is unknown. By measuring only the relative distance using the onboard ranging sensors, the target-center position can be estimated accurately based on a fuzzy wavelet neural network (FWNN) and the least squares fit method.
  \item In contrast to the existing works \cite{deghat2015multi,dong2017time}, we estimate the center position of the targets directly without knowing the real-time position of all targets, which avoids the accumulation of errors caused by estimating the real-time position of each target. In addition, the desired trajectory of the two tasking agents encircling all targets can be quickly obtained by direct estimation of the center position, which implies that the direct estimation of the center position can improve the speed of the whole encirclement algorithm.
  \item Different from the existing encirclement controllers, based on anti-synchronization (AS) theory and the position estimator \cite{Liu2022Bounded,Liu2021Anti}, our designed the encirclement controller can not only be beneficial for the controller and estimator to act simultaneously to ensure the global stability of the system, but also enables the two agents to circumnavigate all targets symmetrically to ensure maximum sensor coverage of all targets.
\end{enumerate}

Notations:
$\bx\in \mathrm{R}^ {n\times m}$ is the ${n\times m}$ real matrix. $I_{n\times n}$ and $1_n$ used to denote the $n$ dimensional identity matrix and identity vector, respectively. $P^T$ is the transpose of the matrix $P$. The eigenvalue of a square matrix is represented as the symbol $\lambda$. $\|\cdot\|$ is the L2 norm, and $|\cdot|$ represents the absolute value. $\mathcal{R}\big\{\cdot\big\}$ represents the rounding function. $<\Upsilon(r(k),k),\bp_{iC}(k)>$ represents the included angle between the vectors $\Upsilon(r(k),k)$ and $\bp_{iC}(k)$.

\section{Problem formulation}
\label{sec:Problem_formulation}
\subsection{System models}
Firstly, the model for the two tasking agents is given by
\begin{equation}\label{eq1-1}
\begin{split}
&\bx_{i}(k+1)=\bx_{i}(k)+\bu_i(k),
\end{split}
\end{equation}
where $\bx_i(k)=[x_i(k),y_i(k),z_i(k)]^T, i\in \Phi_1\triangleq\{1,2\}$ and $\bu_i(k)$ are the position and the controlled input of the agent $i$, respectively. $x_i(k)$, $y_i(k)$ and $z_i(k)$ represent the positions of the agent $i$ in the $\mathbf{X}$, $\mathbf{Y}$ and $\mathbf{Z}$ axes of the corresponding coordinate system at sampling instant $k$.

Since the targets are completely unknown, we assume that the model of the target $j, j \in \Phi_2\triangleq\{1,2, \ldots, M\}$ is
\begin{equation}\label{eq1-2}
\begin{split}
&\bs_j(k+1)=\bs_j(k)+\bh_j(k),\\
\end{split}
\end{equation}
where $\bs_j(k)=[s_{xj}(k),s_{yj}(k),s_{zj}(k)]^T$ is the position of the target $j$. $\bh_j(k)\in \mathrm{R}^3$ represents the unknown displacement of the target $j$ moving trajectory, which can be continuous and nonlinear.

Generally, to encircle the free moving targets, the expected encirclement trajectories of two tasking agents need to be predefined, e.g., a circular trajectory with radius $r(k)$ and center $C(k)=[C_{x}(k),C_{y}(k),C_{z}(k)]^T$. In this work, $C(k)$ is named as the target-center and defined as $C(k)\triangleq\sum^M_{j=1}\gamma_j\bs_j, 0<\gamma_j<1$, where $\sum^M_{j=1}\gamma_j=1$. Based on the model \eqref{eq1-2}, the dynamic of $C(k)$ can be obtained as
\begin{equation}\label{eq1-3}
\begin{split}
C(k+1)=C(k)+\bh(k),
\end{split}
\end{equation}
where $\bh(k)=\sum^M_{j=1}\gamma_j\bh_j(k)$ represents the target-center displacement.

\begin{figure}
\setlength{\abovecaptionskip}{-0.05cm}
\centering
  \includegraphics[width=8cm]{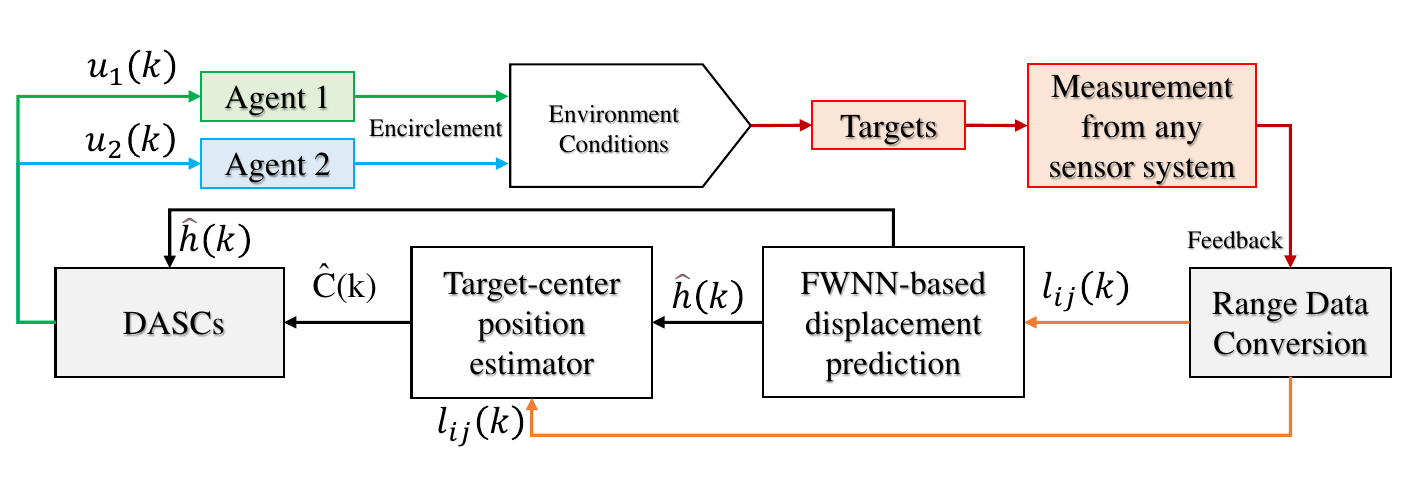}
 \caption{Targets encirclement control and feedback loop.}
  \label{fig:uav_control_feedback_loop}
\end{figure}
\setlength{\belowcaptionskip}{-0.5cm}

\subsection{Objective}
The primary goal of this work is to achieve the tracking and encirclement of multiple targets by using only two mobile tasking agents. To accomplish this, we design the targets encirclement estimate-control loop as shown in Fig. \ref{fig:uav_control_feedback_loop}, which includes the target displacement prediction, the target-center position estimator, and the DASCs. It is worth noting that the positioning and control algorithms can be designed solely based on the distance information measured by the on-board sensors of the two tasking agents.

Therefore, the encirclement of multiple non-cooperative moving targets discussed in this article achieves at least the following goals.

\textbf{Estimator:} The objective of the proposed estimator is to directly estimate the position of the target-center without knowing the positions or models of individual targets. Therefore, the following definitions can be proposed.

\begin{definition}
If the displacement $\bh_j(k)$ of the target is continuous and nonlinear, an ideal FWNN can be designed to predict the displacement $\bh(k)$ of the target-center when the following convergence holds,
\begin{equation}\label{eq1-4}
\begin{split}
\underset{k\rightarrow \infty}{\mathrm{lim}}\{\bh(k)-\hat{\bh}(k)\}=0,
\end{split}
\end{equation}
where $\hat{\bh}(k)$ is the predicted displacement of target-center.
\end{definition}

\begin{definition}
Under the action of the estimator, the two tasking agents can estimate the position of target-center if the following is true,
\begin{equation}\label{eq1-4-2}
\begin{split}
\underset{k\rightarrow \infty}{\mathrm{lim}}\{C(k)-\hat{C}(k)\}=0,
\end{split}
\end{equation}
where $\hat{C}(k)=[\hat{C}_{x}(k),\hat{C}_{y}(k),\hat{C}_{z}(k)]^T$ denotes the target-center position estimator at sampling instant $k$.
\end{definition}

\textbf{Controller:} The goal of the proposed controller is firstly to control the two tasking agents to travel in an AS manner, then secondly, the two agents have to circumnavigate around the center of the target-center with the dynamical radius $r(k)$. Accordingly, we give the following definition.
\begin{definition}
With the action of the controller, the two tasking agents achieve the target-centered AS and the encirclement of all non-cooperative targets if the following conditions hold,
\begin{equation}\label{eq1-5}
\left\{\begin{split}
\underset{k\rightarrow \infty}{\mathrm{lim}}\{\bp_{1C}(k)&+\bp_{2C}(k)\}=0,\\
\underset{k\rightarrow \infty}{\mathrm{lim}}||\bp_{iC}(k)&+\Upsilon(r(k),k)||^2\leq\delta,\\
\end{split}\right.
\end{equation}
where $p_{iC}=\bx_i(k)-C(k), i\in \Phi_1$ is the relative position from Agent $i$ to target-center. $\Upsilon(r(k),k)\in \mathrm{R}^3$ represents the designed encirclement trajectory of two tasking agents, and $\delta$ is the tracking error of the designed encirclement trajectory. Notice that $\delta$ is very small.
\end{definition}
\begin{remark}
In this work, since the motion of each target is independent, the distribution of all targets is variable. As mentioned above, the radius of the designed trajectory will vary to ensure that the two tasking agents can encircle all the targets. When the radius is constant $r$, we can guarantee that the condition $\|\bp_{iC}(k)\|=r$ holds at all the times. However, if the radius is dynamic, we can only guarantee that \eqref{eq1-5} hold. At the moment when the radius changes, the target-center position can be located in the middle point of the line between two tasking agents, but $\|\bp_{iC}(k)\|=r(k)$ is not guaranteed to be true.
\end{remark}

In order to design the estimator and controller reasonably for achieving the encirclement of all non-cooperative targets, the following assumptions have been made.

\begin{assumption}
The maximum velocities of the two agents should be greater than the maximum velocities of all targets, so as to ensure that the two tasking agents can track and circle all targets at the same time.
\end{assumption}

\begin{assumption}
Without loss of generality, the two tasking agents are denoted as Agent 1 and Agent 2. There is direct communication between the two tasking agents, and all agents are considered to be in the same coordinate system from initial stage.
\end{assumption}

\begin{assumption}
To ensure two tasking agents encircle all targets while remaining opposite to one another, the included angle between the preset trajectory $\Upsilon(r(k),k)$ and the relative position $\bp_{iC}(k)$ is set as
\begin{equation}\label{eq1-6}
\left\{\begin{split}
&<\Upsilon(r(k),k),\bp_{1C}(k)>=0^{\circ},\\
&<\Upsilon(r(k),k),\bp_{2C}(k)>=180^{\circ}.
\end{split}\right.
\end{equation}
\end{assumption}

\begin{assumption}
By using the onboard sensors in the two tasking agents, such as vision sensors, LiDAR, etc, the following self-displacement $\bw_i(k)$, and distance $l_{ij}(k)$ from the agent $i$ to the target $j$, can be measured, i.e.,
\begin{equation}\label{eq1-7}
\left\{\begin{split}
&\bw_i(k)=\bx_i(k+1)-\bx_i(k),\\
&l_{ij}(k)=||\bx_i(k)-\bs_j(k)||.
\end{split}\right.
\end{equation}
\end{assumption}

\section{Estimate-control law}
\subsection{Estimator design}
Based on the Assumption 3 and by calculating the distance error $l^2_{1j}(k)-l^2_{2j}(k)$, the following position output variable that is related to the target $j$ can be obtained,
\begin{equation}\label{eq1-8}
\begin{split}
\psi_j(k)=&\bp_{12}^T(k)\bs_j(k)\\
=&-\frac{1}{2}(l_{1j}^2(k)-l_{2j}^2(k)-\bx_1^T(k)\bx_1(k)\\
&+\bx_2^T(k)\bx_2(k)),
\end{split}
\end{equation}
where $\bp_{12}(k)=\bx_1(k)-\bx_2(k)$ is the relative position from Agent $1$ to Agent $2$, and $||\bp_{12}(k)||=\ell_{12}(k)$. $\bx_i(k)$ is the actual measured position of agent $i$, which can be obtained by the accumulation of displacements, that is,  $\bx_1(k)=\bx_1(0)+\sum_{\kappa=0}^{k-1}\bw_1(\kappa)$ and $\bx_2(k)=\bx_2(0)+\sum_{\kappa=0}^{k-1}\bw_2(\kappa)$.

\begin{remark}
One may argue that the position can be found by accumulating the controller output $\bu_i(k)$. In reality, there would always be systematic bias and random walk-based noises between the controller output and the actual displacement, thus the active correction by real-time feedback systems is required.
\end{remark}

Therefore, the target-center position output variable is defined as
\begin{equation}\label{eq1-9}
\begin{split}
\psi(k)=\sum^M_{j=1}\gamma_j\psi_j(k).
\end{split}
\end{equation}

Based on the variables obtained above, the positions of all targets can not be obtained directly. In order to ensure that all targets can be encircled, an estimator for the center of the targets is developed by avoiding estimating the positions of all targets, which is different from the existing works \cite{matveev2017range,deghat2015multi}.
Note that the precondition of an accurate target-center estimation by the least square fit is that the target model should be known. Since the target models are unknown in this work, a FWNN is chosen to predict the displacement of the target-center. Based on this model, the position of target-center can be further estimated.

Denote the actual displacement output variable as $\Delta\psi(k)=\psi(k+1)-\psi(k)\simeq\bp_{12}^T(k)\bh(k)$. Furthermore, $\Delta\psi(k)$ is mapped to the corresponding fuzzy set $\{d_{\iota1}, \ldots, d_{\iota\nu}\}$. Then, $\bh(k)$ can be approximated by the FWNN and reasoning through the following fuzzy rules.

Plant Rule $\iota, \iota \in \Phi_\iota = \{1, \ldots, l\}$ : If $\theta_{1}(k)$ is $d_{\iota1}$, $\theta_{2}(k)$ is $d_{\iota2}$, \ldots, $\theta_{\nu}(k)$ is $d_{\iota \nu}$, then $\bh(k)$ is $\Lambda_{\iota}$.

In these rules, $l$ is the number of fuzzy rules and $\nu$ is the size of fuzzy sets. $\theta_{1}(k), \ldots, \theta_{\nu}(k)$ are the input variables, which can be designed later. $\Lambda_{\iota}$ is the fuzzy output.

By mathematically describing the above fuzzy rules, the fuzzy basis function is given as
\begin{equation*}
\begin{split}
&\vartheta_{\iota}(k)=\frac{\prod_{\imath=1}^{\nu}\tilde{d}_{\iota\imath }(\theta_{\imath}(k))}{\sum^{l}_{\iota=1}\prod_{\imath=1}^{\nu}\tilde{d}_{\iota\imath }(\theta_{\imath}(k))}
\end{split}
\end{equation*}
with $\sum^{l}_{\iota=1}\vartheta_{\iota}(k)=1$. $\tilde{d}_{\iota\imath }(\theta_{\imath}(k))$ is the membership grade of $\theta_{\imath}(k)$ in $d_{\iota\imath},\imath\in \Phi_\nu=\{1,2,\ldots,\nu\}$. According to the Gaussian function, the membership grade $\tilde{d}_{\iota\imath}(\theta_{\imath}(k))$ can be described as
\begin{equation*}
\begin{split}
\tilde{d}_{\iota\imath }(\theta_{\imath}(k))&=\exp\left\{\frac{-\theta_{\imath}^2(k)}{2\varsigma^2}\right\},
\end{split}
\end{equation*}
and the input variable is defined as $\theta_{\imath}(k)=\Delta\psi(k)-\sigma+\varrho\imath$, where $(\sigma-\varrho\iota)$ and $\varsigma$ are the given mean and variance of the Gaussian function, respectively. To reflect the generality, the variance $\varsigma$ will be designed to be large enough to cover all movements of the target.

\begin{figure}
\centering
  \includegraphics[width=7cm]{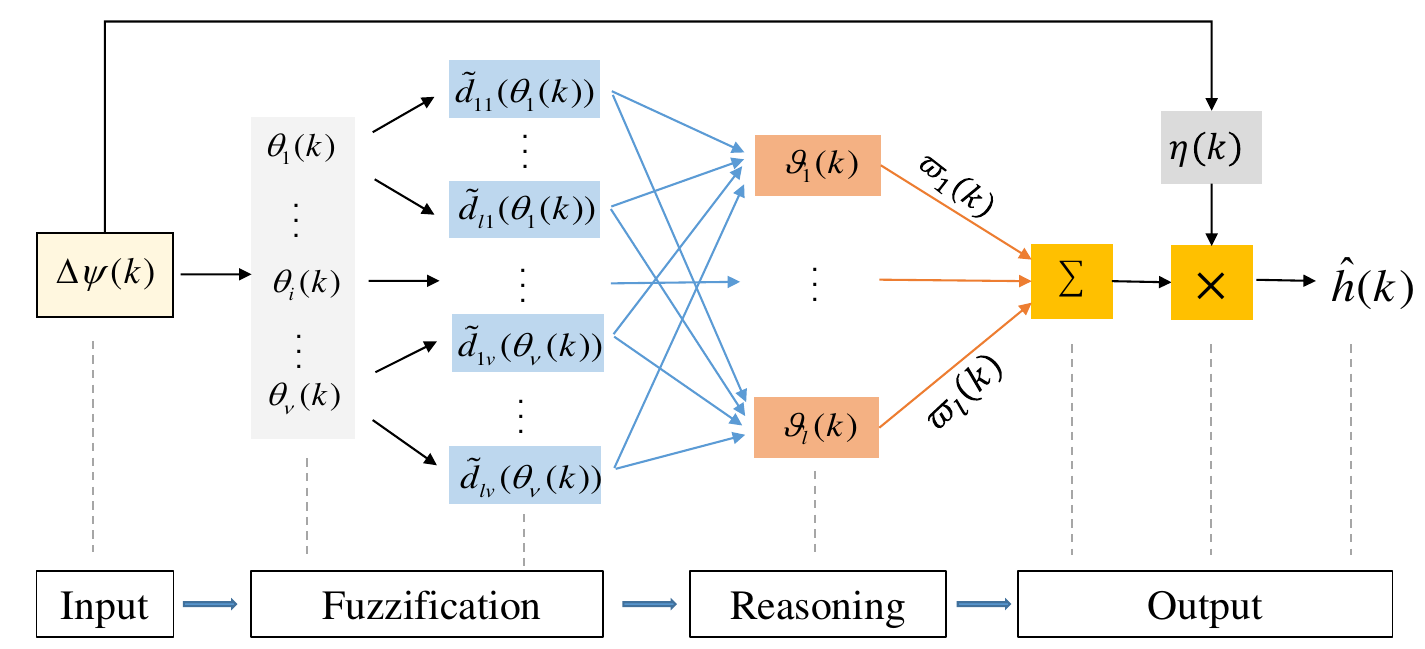}
 \caption{The structure of FWNN.}
  \label{NN}
\end{figure}

Corresponding to the reasoning process mentioned above, the FWNN contains six layers, as shown in Fig. \ref{NN}. The first layer is the input layer. The second and third layers are fuzzification layers consisting of the membership grade functions $\tilde{d}_{\iota\imath }(\theta_{\imath}(k))$. The fourth layer is a reasoning layer based on the fuzzy basis function $\vartheta_{\iota}(k)$, and the fifth and sixth layers are output layers that are defuzzified and then wavelet-processed by the reasoning results. Moreover, the actual displacement output variable $\Delta\psi(k)$ is the input of the network. The number of neurons in the second, third and fourth layers are $\nu$, $l\nu$ and $l$ respectively. $\eta(k)$ is a mother wavelet function, and satisfies $\int^{+\infty}_{-\infty}
|\eta(k)|^2=1$. $\varpi_{\iota}(k)\in \mathrm{R}^3$ is the adaptive updated weight.

Recalling the variable in equation \eqref{eq1-9}, the following value function of FWNN can be denoted as
\begin{equation}\label{eq1-10}
\begin{split}
J(k)=&\frac{1}{2}\Big\{\Delta\psi(k)-\bp_{12}^T(k)\hat{\bh}(k)\Big\}^2,\\
\end{split}
\end{equation}
where $\hat{\bh}(k)$ is the output of the FWNN.

Then, the appropriate output weight needs to be designed so as to minimize the value function $J(k)$, that is
\begin{equation}\label{eq1-11}
\begin{split}
\varpi_{\iota}(k)=arc\min\{J(k)\}\in \mathrm{R}^3.
\end{split}
\end{equation}

Based on the gradient descent method, the updating law of the output weight can be designed as
\begin{equation}\label{eq1-12}
\begin{split}
\varpi_{\iota}(k+1)&=\varpi_{\iota}(k)+\triangle\varpi_{\iota}(k)\\
&=\varpi_{\iota}(k)+\alpha \bp_{12}^T(k)\hat{\be}_h(k)\bp_{12}(k)\eta(k)\vartheta_{\iota}(k),
\end{split}
\end{equation}
where $\triangle\varpi_{j\iota}(k)=-\alpha\frac{\partial J(k)}{\partial \varpi_{\iota}(k)}$, and $\alpha \in (0,1)$ is the learning rate that need to be designed. $\hat{\be}_h(k)=\bh(k)-\hat{\bh}(k)$ is the displacement  prediction error of the target-center by FWNN.

To summarize, the predicted displacement of the target-center can be rewritten as
\begin{equation}\label{eq1-13}
\begin{split}
\hat{\bh}(k)=\sum^l_{\iota=1}\eta(k)\varpi_{\iota}(k)\vartheta_{\iota}(k).
\end{split}
\end{equation}

Thus, the dynamic of the target-center position estimator can be designed as
\begin{equation}\label{eq1-14}
\left\{\begin{split}
\hat{C}(k+1)=&\hat{C}(k)+\hat{\bh}(k)+K_s(k+1)\be_{out}(k+1),\\
\hat{\psi}(k)=&\bp_{12}^T(k)(\hat{C}(k-1)+\hat{\bh}(k-1)),
\end{split}\right.
\end{equation}
where $\hat{\psi}(k)$ are the prediction of $\psi(k)$. Denoted $\be_{out}(k)=\psi(k)-\hat{\psi}(k)$, the part $K_s(k+1)\be_{out}(k+1)$ is used to adjust the estimated position, where $K_s(k) \in \mathrm{R}^{3\times3}$ is the gain of the estimator. Based on the least-squares fit in \cite{johnstone1982exponential}, $K_s(k)$ can be designed as the following formula,
\begin{equation}\label{eq1-15-1}
\begin{split}
K_s(k)&=\frac{\xi_s(k-1)\bp_{12}(k)}{\vartheta_1\vartheta_2+\bp_{12}^T(k)\xi_s(k-1)\bp_{12}(k)},\\
\end{split}
\end{equation}
where the covariance matrix $\xi_s(k) \in \mathrm{R}^{3\times 3} (\xi_s(k)>0)$ is defined by
\begin{equation}\label{eq1-15-2}
\begin{split}
\xi_s^{-1}(k)=&\vartheta_1\xi_s^{-1}(k-1)+\frac{1}{\vartheta_2}\bp_{12}(k)\bp_{12}^T(k)
\end{split}
\end{equation}
with the exponential forgetting factor $\vartheta_1 \in [0,1]$ and the new information utilization factor $\vartheta_2 \in [0,1]$.

Based on Definition 1, the target-center position estimation error is defined as
\begin{equation}\label{eq1-16}
\begin{split}
\hat{\be}_s(k)=C(k)-\hat{C}(k).
\end{split}
\end{equation}

Recalling the model \eqref{eq1-3} and the estimator \eqref{eq1-14}, the dynamic of the target-center
position estimation error can be further obtained as
\begin{equation}\label{eq1-17}
\begin{split}
\hat{\be}_s(k+1)=(I_{n\times n}-K_s(k+1)\bp_{12}^T(k+1))(\hat{\be}_s(k)+\hat{\be}_h(k)).
\end{split}
\end{equation}

\vspace{-0.8cm}
\subsection{Controller design}
\begin{figure}
\setlength{\abovecaptionskip}{-0.05cm}
\centering
  \includegraphics[width=7cm]{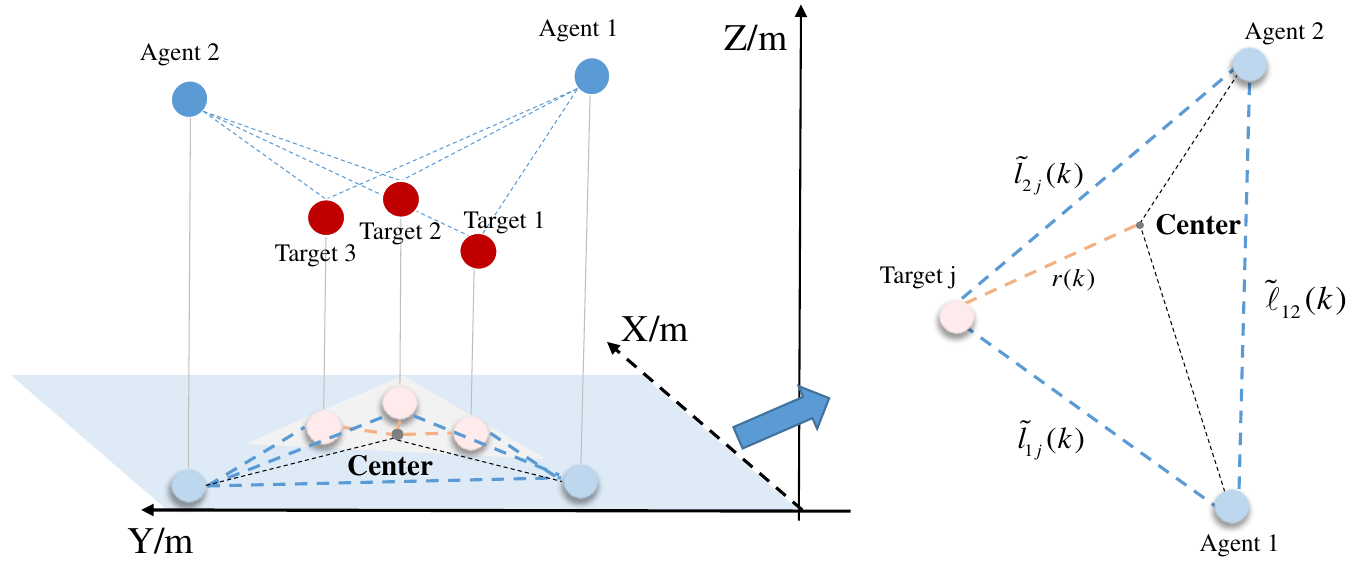}
 \caption{3D positions are
orthogonally projected on the 2D ground plane.}
  \label{radius}
\end{figure}

Define $\hat{p}_{i\hat{C}}(k)=\bx_i(k)-\hat{C}(k)$ as the relative position from the tasking agents to the estimated target-center. Based on the target-center position estimator \eqref{eq1-14} and the Assumption 3, the DASCs are designed as
\begin{equation}\label{eq1-18}
\left\{\begin{split}
\bu_1(k)=&\beta\big\{\hat{\bp}_{1\hat{C}}(k)+\Upsilon(r(k),k)\big\}+\hat{\bh}(k),\\
\bu_2(k)=&\beta\big\{\hat{\bp}_{2\hat{C}}(k)-\Upsilon(r(k),k)\big\}+\hat{\bh}(k),
\end{split}\right.
\end{equation}
where $\beta \in R^1$ is the gain of controllers, which will be designed later.
In order to make the two agents be able to encircle all moving targets in a smooth circle, the preset trajectory $\Upsilon(r(k),k)$ is defined as
\begin{equation*}
\begin{split}
\Upsilon(r(k),k)=\left[
                \begin{array}{c}
                  r(k)sin(\rho k\pi)cos(\tilde{\rho}(k)) \\
                  r(k)cos(\rho k\pi) \\
                  r(k)cos(\rho k\pi)sin(\tilde{\rho}(k)) \\
                \end{array}
              \right]
\end{split}
\end{equation*}
where $0<\rho<1$ denotes the frequency of circumnavigation, which has to be small enough so that the condition $<\Upsilon(r(k),k),\bp_{1C}(k)>\simeq<\Upsilon(r(k-1),k-1),\bp_{1C}(k)>$ holds. The elevation angle is defined as $\tilde{\rho}(k)= arctan2{\frac{z_1(k)-\hat{C}_z(k)}{\sqrt{(x_1(k)-\hat{C}_x(k))^2+(y_1(k)-\hat{C}_y(k))^2}}}$. $r(k)$ is the radius of the two agents' trajectories. As shown in Fig. \ref{radius}, 3D positions are orthogonally projected onto the 2D ground plane, and $r(k)$ can be found as the following formulas by founding the distance of projected 2D point.
\begin{equation}\label{eq1-19}
\begin{split}
r(k)=&\mathcal{R}\big\{\max_{j\in\Phi_2}\{\sqrt{\tilde{l}_{2j}^2(k)+\hat{l}_{2\hat{C}}^2(k)-2\hat{l}_{\bar{\theta}}}\}\big\}+b,\\
\hat{l}_{\bar{\theta}}=&\tilde{l}_{2j}(k)\hat{l}_{2\hat{C}}(k)\cos(\bar{\theta}_{\hat{C}1}(k)-\bar{\theta}_{j1}(k))\\
\bar{\theta}_{\hat{C}1}(k)=&\emph{acos} \left\{ \frac{\hat{l}^2_{2\hat{C}}(k)+\tilde{\ell}^2_{12}(k)-\hat{l}^2_{1\hat{C}}(k)}{2\hat{l}_{2\hat{C}}(k)\tilde{\ell}_{12}(k)}\right\},\\
\bar{\theta}_{j1}(k)=&\emph{acos}\left\{\frac{\tilde{l}^2_{2j}(k)+\tilde{\ell}^2_{12}(k)-\tilde{l}_{1j}^2(k)}{2\tilde{l}_{2j}(k)\tilde{\ell}_{12}(k)}\right\},
\end{split}
\end{equation}
where
\begin{equation*}
\begin{split}
\hat{l}_{i\hat{C}}(k)&=\sqrt{(x_i(k)-\hat{C}_{x}(k))^2+(y_i(k)-\hat{C}_{y}(k))^2},\\
\tilde{\ell}_{12}(k)&=\sqrt{(x_1(k)-x_2(k))^2+(y_1(k)-y_2(k))^2},\\
\tilde{l}_{ij}(k)&\simeq\sqrt{l^2_{ij}(k)-(z_2(k)-\hat{C}_{z}(k))^2},
\end{split}
\end{equation*}
 and $b>0$ is a given constant that makes the radius of the preset trajectory large enough so as to avoid collision between the two agents and all targets. $\emph{acos}$ is the inverse coine function. Suppose that the maximal value of $r(k)$ is $r$, where $r>b$. Then, we have $b\leq\|\Upsilon(r(k),k)\|\leq r$.
\setlength{\belowcaptionskip}{-0.5cm}
\begin{remark}
 In this work, we do not consider active obstacle avoidance issues because that requires wide-angle sensor coverage for each target. We perform obstacle avoidance passively by planning the AS controller according to the observation feedback so that the controlled agents do not collide with each other. Moreover, the dynamical radius $r(k)$ is designed by considering the maximum distance between all targets and the target-center. The parameter $b$ is devised to prevent any collision between the two agents and all targets. Therefore, the controller is robust enough to prevent any collision from happening when all feedback systems are running normally.
 \end{remark}

Based on Definition 3, the AS errors can be denoted by $\be(k)=\bp_{1C}(k)+\bp_{2C}(k)$, $\be_1(k)=\bp_{1C}(k)+\Upsilon(r(k),k)$ and $\be_2(k)=\bp_{2C}(k)-\Upsilon(r(k),k)$. Then, considering the models \eqref{eq1-1}, \eqref{eq1-3}, and the condition $<\Upsilon(r(k+1),k+1),\bp_{1C}(k)>\simeq<\Upsilon(r(k),k),\bp_{1C}(k)>$, the dynamic of the AS errors can be derived as
\begin{equation}\label{eq1-20-1}
\left\{\begin{split}
&\be(k+1)=(1+\beta)\be(k)+2\beta\hat{\be}_s(k)-2\hat{\be}_h(k),\\
&\be_1^T(k+1)\be_1(k+1)\simeq\left\{\begin{split}
&\tilde{\be}_1^T(k)\tilde{\be}_1(k), k\in \hat{k},\\
&\tilde{\be}_1^T(k)\tilde{\be}_1(k)+\delta_1(k+1),k \notin \hat{k},\\
\end{split}\right.\\
&\be_2^T(k+1)\be_2(k+1)\simeq\left\{\begin{split}
&\tilde{\be}_2^T(k)\tilde{\be}_2(k), k\in \hat{k},\\
&\tilde{\be}_2^T(k)\tilde{\be}_2(k)+\delta_2(k+1),k \notin \hat{k},\\
\end{split}\right.
\end{split}\right.
\end{equation}
where $\tilde{\be}_i(k)=(1+\beta)\be_i(k)+\beta\hat{\be}_s(k)-\hat{\be}_h(k)$ and $\delta_i(k+1)=\big(2\|\bp_{iC}(k+1)\|+r(k+1)+r(k)\big)(r(k+1)-r(k))$. $\hat{k}\triangleq\{k_0,k_1,k_2,\ldots,k_\infty\}$ is the sampling sequence set with the initial value $k_0=0$, and the sampling interval is defined as $t_\jmath=k_{\jmath+1}-k_{\jmath}, \jmath\in\{1,2,\ldots,\infty\}$. $k\in\hat{k}$ for $r(k)=r(k+1)$, else $k\notin \hat{k}$.

Recalling the DASCs in \eqref{eq1-18}, the formulation $\bu_{12}(k)=\bu_{1}(k)-\bu_{2}(k)=\beta\big\{\bp_{12}(k)+2\Upsilon(r(k),k)\big\}$ can be obtained. Since $\bp_{12}(k+1)=\bp_{12}(k)+\bu_{12}(k)$, we have $\|\bu_{12}(k)\|\leq\bar{\mu}$ with $\bar{\mu}=|\beta|(\ell_{12}(0)+2r)$ for $0<|1+\beta|<1$. Based on this, $\bar{\nu}$ satisfying $0\leq\lambda\{\bp_{ij}(k)\bp_{ij}^T(k)\}\leq\bar{\nu}$ can be estimated.

Then, the following lemma is given and will be useful for the convergence analysis in the next section.

\begin{lemma}
The sequence $\{\bp_{12}(k)\}, k\in[\kappa,\kappa+N-1], \forall \kappa\in Z, \exists N \in Z^{+}$ is persistently exciting if the following inequality holds for $||\bu_{12}(k)||\leq\bar{\mu}$,
\begin{equation}\label{eq1-20-2}
\begin{split}
0<\hat{a}I_{n\times n}\leq\sum_{k=\kappa}^{\kappa+N-1}\bp_{12}(k) \bp_{12}^T(k)\leq\check{a}I_{n\times n}<\infty,
\end{split}
\end{equation}
where $N$ is the motion period that occurs when the two tasking agents are moving around all targets, and
\begin{equation*}
\begin{split}
\check{a}=&N\ell_{12}^2(\kappa)
+N(N-1)\ell_{12}(\kappa)\bar{\mu}\\&+\frac{1}{6}N(N-1)(2N-1)\bar{\mu}^2,\\
\hat{a}=&N\ell_{12}^2(\kappa)
-N(N-1)\ell_{12}(\kappa)\bar{\mu}\\
&+\frac{1}{6}N(N-1)(2N-1)\bar{\mu}^2.
\end{split}
\end{equation*}
\end{lemma}
Proof:
Based on $\bp_{12}(k+1)=\bp_{12}(k)+\bu_{12}(k)$, the following equality can be obtained $\forall k\in Z$,
\begin{equation}\label{eq1-21}
\begin{split}
1_n^T\bp_{12}(k+1)\bp_{12}^T(k+1)1_n=(\bp_{12}^T(k)1_n&+\bu_{12}^T(k)1_n)^2.
\end{split}
\end{equation}

Furthermore, considering the conditions $\|\bu_{12}^T(k)1_n\|\leq \bar{\mu}$, the following inequality is derived,
\begin{equation}\label{eq1-22}
\begin{split}
(\ell_{12}(k)-\bar{\mu})^2&\leq 1_n^T\bp_{12}(k+1)\bp_{12}^T(k+1)1_n\\\leq(\ell_{12}(k)&+\bar{\mu})^2,~~~~~~\forall k\in Z.
\end{split}
\end{equation}

Then, considering the motion period $k\in[\kappa,\kappa+N-1]$, there always exists an
$N \in Z^{+}$ such that the persistently exciting condition of $\{\bp_{12}(k)\}$ can be obtained as the inequality \eqref{eq1-20-2}.

\section{Convergence analysis}
In this section, it is first proven that FWNN can approximate the model of the target-center. Next, the convergence of the estimation error and the AS error can be analyzed.

\begin{theorem}
The FWNN can predict the displacement of the target-center position with a learning rate $\alpha$, which satisfies the following condition,
\begin{equation}\label{eq1-23}
\begin{split}
&0<\alpha<\frac{2}{\overline{\nu}}.
\end{split}
\end{equation}
\end{theorem}
Proof:
In order to analyze the convergence of $\hat{\be}_h(k)$, the following Lyapunov candidate function (LCF) is chosen,
\begin{equation}\label{eq1-24}
\begin{split}
V_1(k)=\frac{1}{2}\hat{\be}_h^T(k)\hat{\be}_h(k).
\end{split}
\end{equation}

According to the literature \cite{ku1995diagonal}, the difference in the LCF can be obtained as
\begin{equation}\label{eq1-25}
\begin{split}
\triangle V_1(k)=&V_1(k+1)-V_1(k)\\
=&\triangle\hat{\be}_h^T(k)\{\hat{\be}_h(k)+\frac{1}{2}\triangle\hat{\be}_h(k)\},
\end{split}
\end{equation}
where
\begin{equation*}
\begin{split}
\triangle\hat{\be}_h(k)=\hat{\be}_h(k+1)-\hat{\be}_h(k)=\sum^l_{\iota=1}\frac{\partial\hat{\be}_h(k)}{\partial\varpi_{\iota}(k)}\triangle\varpi_{\iota}(k).\\
\end{split}
\end{equation*}

Recalling the formula \eqref{eq1-12}, $\triangle\varpi_{j\iota}(k)$ can be reacquired as
\begin{equation}\label{eq1-26}
\begin{split}
\triangle\varpi_{\iota}(k)=-\alpha \bp_{12}^T(k)\hat{\be}_h(k)\bp_{12}(k)\frac{\partial\hat{\be}_h(k)}{\partial\varpi_{\iota}(k)}.
\end{split}
\end{equation}


Denoting $\lambda\{\bp_{12}(k)\bp_{12}^T(k)\}=\nu$ and calculating $\frac{\partial\hat{\be}_h(k)}{\partial\varpi_{\iota}(k)}=-\eta(k)\vartheta_{\iota}(k)$, we have
\begin{equation}\label{eq1-28}
\begin{split}
\triangle V_1(k)\leq&-\alpha\nu\eta^2(k)\sum^l_{\iota=1}\vartheta_{\iota}^2(k)\hat{\be}_h^T(k)\hat{\be}_h(k)\\
&+\frac{1}{2}\Big\{\alpha \nu\eta^2(k)\sum^l_{\iota=1}\vartheta_{\iota}^2(k)\Big\}^2\hat{\be}_h^T(k)\hat{\be}_h(k)\\
=&-\alpha(2\psi-\alpha\psi^2) V_1(k),
\end{split}
\end{equation}
where $\psi=\nu\eta^2(k)\sum^l_{\iota=1}\vartheta_{\iota}^2(k)$.

To let the estimation error converge exponentially, it is only necessary to ensure that both $0<\alpha<1$ and $0<2\psi-\alpha\psi^2<1$ hold. $\psi\leq\bar{\nu}$ as a result from $\sum^l_{\iota=1}\vartheta_{\iota}(k)=1$, $\int^{+\infty}_{-\infty}
|\eta(k)|^2=1$, and $\nu\leq\bar{\nu}$. Furthermore, under the condition \eqref{eq1-23} , $0<\alpha(2\psi-\alpha\psi^2)<1$ can be obtained.

Therefore, the prediction error of FWNN converges exponentially to zero. Based on Definition 1, the FWNN is regarded as can predict the displacement of the virtual target-center position.

\begin{theorem}
The estimator \eqref{eq1-14} can estimate the virtual target-center position by selecting appropriate factor $\vartheta_1 \in (0,\frac{1}{2})$.
\end{theorem}
Proof:
Recalling the formula \eqref{eq1-15-2} and Lemma 1, the upper and lower bounds of the mean error $\xi_s(k)$ can easily be deduced as follows:
\begin{equation}\label{eq1-29}
\left\{\begin{split}
&\xi_0^{-1}(k)\geq\hat{b},~\forall k\geq N-1,\\
&\xi_0^{-1}(k)\leq\check{b}_{k},~\forall k\in Z,\\
\end{split}\right.
\end{equation}
where
\begin{equation*}
\begin{split}
\hat{b}=&\frac{N(1-\frac{1}{\vartheta_1})}{\vartheta_2(1-\frac{1}{\vartheta_1^N})}\hat{a}I_{n\times n},\\
\check{b}_{k}=&\frac{(1-\vartheta_1)\vartheta_1^k}{1-\vartheta_1^N}\sum_{\ell=0}^{N-1}\xi_0^{-1}(\ell)+\frac{(1-\vartheta_1^k)N}{(1-\vartheta_1^N)\vartheta_2}\check{a}I_{n\times n}.
\end{split}
\end{equation*}

Considering the target-center position estimator error $\hat{\be}_s(k+1)$ in \eqref{eq1-17}, we choose the following LCF,
\begin{equation}\label{eq1-30}
\begin{split}
V_2(k)=\frac{1}{2}\hat{\be}_s^T(k)\xi_s^{-1}(k)\hat{\be}_s(k).
\end{split}
\end{equation}

Next, the difference in the LCF can be deducted by
\begin{equation}\label{eq1-31}
\begin{split}
\triangle V_2(k)=&\frac{1}{2}(\hat{\be}_s(k)+\hat{\be}_h(k))^T(I_{n\times n}\\
&-K_s(k+1)\bp_{12}^T(k+1))^T\xi_s^{-1}(k+1)\\
&\times (I_{n\times n}-K_s(k+1)\bp_{12}^T(k+1))(\hat{\be}_s(k)\\
&+\hat{\be}_h(k))-\frac{1}{2}\hat{\be}_s^T(k)\xi_s^{-1}(k)\hat{\be}_s(k).
\end{split}
\end{equation}

Based on the formulas \eqref{eq1-15-1}, \eqref{eq1-15-2} and the matrix inversion lemma, the following formulations can be obtained
\begin{equation}\label{eq1-32}
\begin{split}
I_{n\times n}-K_s(k)\bp_{12}^T(k)&=\vartheta_1\xi_s(k)\xi_s^{-1}(k-1).\\
\end{split}
\end{equation}

Then, the formula \eqref{eq1-31} can be further rewritten as
\begin{equation}\label{eq1-33}
\begin{split}
\triangle V_2(k)=&\frac{1}{2}\vartheta_1(\hat{\be}_s(k)+\hat{\be}_h(k))^T\xi_s^{-1}(k))\xi_s(k+1)\\
&\times\xi_s^{-1}(k+1)(I_{n\times n}-K_s(k+1)\bp_{12}^T(k+1))\\
&\times(\hat{\be}_s(k)+\hat{\be}_h(k))-\frac{1}{2}\hat{\be}_s^T(k)\xi_s^{-1}(k)\hat{\be}_s(k)\\
\leq&(2\vartheta_1-1)V_2(k)+\vartheta_1\hat{\be}_h^T(k)\xi_s^{-1}(k))\hat{\be}_h(k).
\end{split}
\end{equation}

Recalling the Theorem 1, for $k\rightarrow\infty$, $||\hat{\be}_h(k)||^2\rightarrow0$ can be obtained. Based on \eqref{eq1-33}, we have
\begin{equation}\label{eq1-34}
\begin{split}
||\hat{\be}_s(k+1)||^2\leq&\frac{(2\vartheta_1)^{k+1}||\check{b}_0||}{||\hat{b}||}||\hat{\be}_s(0)||^2.
\end{split}
\end{equation}

Then, considering the condition $0<2\vartheta_1<1$, the formula \eqref{eq1-4-2} can be further obtained. Based on Definition 2, the estimator \eqref{eq1-14} can estimate the virtual target-center position.

\begin{theorem}
By using the target-center position estimator \eqref{eq1-14} and the DASCs \eqref{eq1-18}, Agent 1 and Agent 2 can achieve AS and
circumnavigate all targets if appropriate controller gain $\beta$ are selected according to the following condition,
\begin{equation}\label{eq1-35}
\begin{split}
-\frac{1}{\sqrt{3}}-1<\beta\leq\frac{1}{\sqrt{3}}-1.
\end{split}
\end{equation}
\end{theorem}
Proof:
According to the AS error \eqref{eq1-20-1}, the LCF can be chosen as
\begin{equation}\label{eq1-36}
\left\{\begin{split}
 V_3(k)=&\frac{1}{3}\be^T(k)\be(k),\\
V_{3i}(k)=&\frac{1}{3}\be_i^T(k)\be_i(k).
\end{split}\right.
\end{equation}

Then, the differences of the LCF can be defined as
\begin{equation}\label{eq1-37}
\left\{\begin{split}
\Delta V_3(k)=&\frac{1}{3}\be^T(k+1)\be(k+1)-\frac{1}{3}\be^T(k)\be(k),\\
\Delta V_{3i}(k)=&\frac{1}{3}\be_{i}^T(k+1)\be_{i}(k+1)-\frac{1}{3}\be_{i}^T(k)\be_{i}(k).
\end{split}\right.
\end{equation}

Next, based on the Cauchy-Schwarz inequality, the differences of the LCF can be further rewritten as
\begin{equation}\label{eq1-38}
\begin{split}
\Delta V_3(k)
\leq& ((1+\beta)^2-\frac{1}{3})\be^T(k)\be(k)+4\beta^2\hat{\be}^T_s(k)\hat{\be}_s(k)\\
&+4\be_h^T(k)\be_h(k),\\
\end{split}
\end{equation}
and
\begin{equation}\label{eq1-39}
\Delta V_{3i}(k)
\leq\left\{\begin{split}
&((1+\beta)^2-\frac{1}{3})\be_i^T(k)\be_i(k)+\beta^2\hat{\be}^T_s(k)\hat{\be}_s(k)\\
&+\be_h^T(k)\be_h(k),k \in \hat{k}\\
&((1+\beta)^2-\frac{1}{3})\be_i^T(k)\be_i(k)+\beta^2\hat{\be}^T_s(k)\hat{\be}_s(k)\\
&+\be_h^T(k)\be_h(k)+\delta_i(k+1),k \notin \hat{k}\\
\end{split}\right.
\end{equation}

Recalling the conclusions mentioned in Theorem 1 and Theorem 2, $||\be_h(k)||^2\rightarrow0$ and $||\hat{\be}_s(k)||^2\rightarrow0$ for $k\rightarrow\infty$ can be obtained. Then, we have
\begin{equation}\label{eq1-40}
\begin{split}
\lim_{k\rightarrow\infty}||\be(k+1)||^2\leq (3(1+\beta)^2)^{k+1}||\be(0)||^2,
\end{split}
\end{equation}
and
\begin{equation}\label{eq1-41}
\begin{split}
\lim_{k\rightarrow\infty}||\be_i(k+1)||^2\leq (3(1+\beta)^2)^{k+1}||\be_i(0)||^2+\hat{\delta},
\end{split}
\end{equation}
where $\hat{\delta}=\sum_{\jmath=0}^{\infty}(3(1+\beta)^2)^{\sum_{t=t_\jmath}^{t_\infty}t}2\big(\sqrt{\overline{\nu}}+r\big)(r-b)\geq\sum_{\jmath=0}^{\infty}(3(1+\beta)^2)^{\sum_{t=t_\jmath}^{t_\infty}t}\delta_i(k_\jmath)$.

Considering the condition \eqref{eq1-35}, the following result is established,
\begin{equation}\label{eq1-42}
\left\{\begin{split}
&\lim_{k\rightarrow\infty}||\be(k+1)||^2\leq0,\\
&\lim_{k\rightarrow\infty}||\be_i(k+1)||^2\leq\delta,\\
\end{split}\right.
\end{equation}
where $\delta=(3(1+\beta)^2)^{t_\infty}2\big(\sqrt{\overline{\nu}}+r\big)(r-b)$. Based on Definition 3, it is expected that the Agent 1 and Agent 2 can
circumnavigate all targets while maintaining AS for maximize sensor coverage.

\section{Numerical simulation and flight experiment}
In this section, firstly, a numerical simulation example in Matlab is used to verify the effectiveness of the proposed target-center estimator and DASC. Secondly, a UAV experimental platform is used to further validate the effectiveness of the proposed algorithm in a real-world environment.

\begin{figure}
\setlength{\abovecaptionskip}{-0.05cm}
\centering
  \includegraphics[width=7cm]{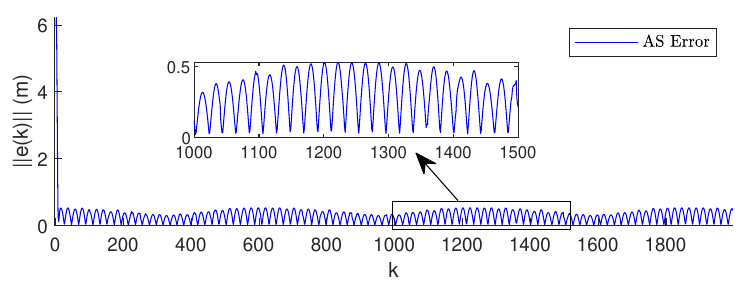}
 \caption{The AS error in the whole simulation process.}
  \label{error}
\end{figure}
\setlength{\belowcaptionskip}{-0.5cm}
\textbf{Numerical simulation:} There exist three targets, i.e, $M=3$, and two agents in the encirclement problem. In order to make the simulation results more intuitive, we assume that all agents and targets are at the same height. Assume the initial positions of all agents, targets and the target-center position estimator in the metric coordinate system as shown below,
\begin{equation*}
\begin{split}
&\bx_{1}(0)=[0,3,0.5]^T,~\bx_{2}(0)=[0,6,0.5]^T,\\
&\bs_{1}(0)=[1,0,0.5]^T,~\bs_{2}(0)=[2,0,0.5]^T,\\
&\bs_{3}(0)=[3,0,0.5]^T,~\hat{C}(0)=[0,0,0.5]^T.
\end{split}
\end{equation*}


The center position of multiple targets is defined based on the concept of geometric center, that is, $\gamma_1=\gamma_2=\gamma_3=\frac{1}{3}$. In the FWNN, we set $l=1$, $\vartheta=5$, $\varsigma=128$, $\sigma=9$, and $\varrho=3$. The mother wavelet function is designed as $\eta(k)=-(0.001\Delta\psi(k)-0.001)\exp(-0.5(0.001\Delta\psi(k)-0.001)^2)$.


\begin{figure}
\setlength{\abovecaptionskip}{-0.05cm}
\centering
  \includegraphics[width=7cm]{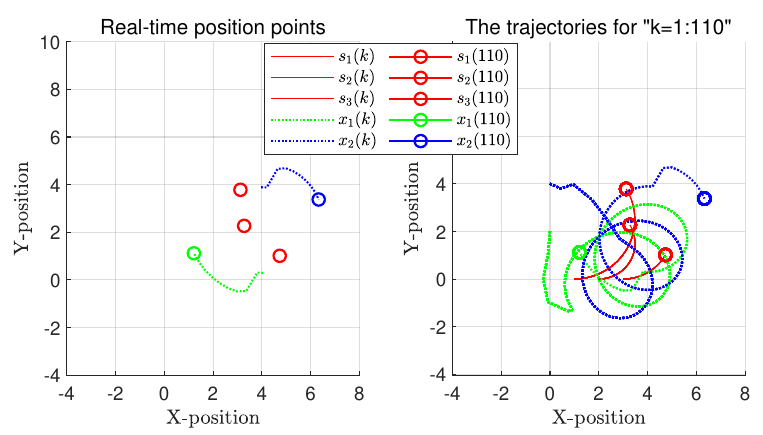}
 \caption{The real-time positions and trajectories at k=110 (times).}
  \label{tra110s}
\end{figure}
\setlength{\belowcaptionskip}{-0.5cm}

\begin{figure}
\setlength{\abovecaptionskip}{-0.05cm}
\centering
  \includegraphics[width=7cm]{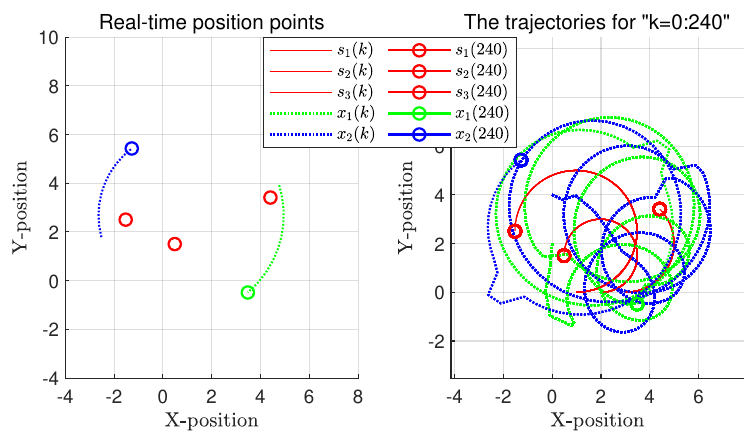}
 \caption{The real-time positions and trajectories at k=240 (times).}
  \label{tra240s}
\end{figure}
\setlength{\belowcaptionskip}{-0.5cm}
In terms of Theorem 2 and Theorem 3, the exponential forgetting factor is designed as $\vartheta_1=0.1$, the new information utilization factor as $\vartheta_2=0.95$ and the controller gain as $\beta=-0.85$. Furthermore, the upper of $\|\bu_1(k)-\bu_2(k)\|$ is derived as $\overline{\mu}=1.8$. Therefore, $\bar{\nu}$ can be obtained as $5.8726$. Based on Theorem 1, the learning rate $\alpha$ is set as $0.01$. The frequency
of circumnavigation is given as $\rho=\frac{1}{24}$, and the radius of preset trajectory $r(k)$ can be obtained based on \eqref{eq1-17} for $b=0.8$.




 Under the action of the controller, the AS error of the two agents is less than 0.5, as shown in Fig. \ref{error}. The real-time position and trajectory maps of all the agents and targets at two moments are given in Fig. \ref{tra110s} and Fig. \ref{tra240s}, respectively. The simulation results show that the two agents are able to encircle all targets in real-time through the dynamic trajectory adjustment under the designed target-center position estimator and the designed DASC.

\begin{figure}
\setlength{\abovecaptionskip}{-0.05cm}
\centering
  \includegraphics[width=8cm]{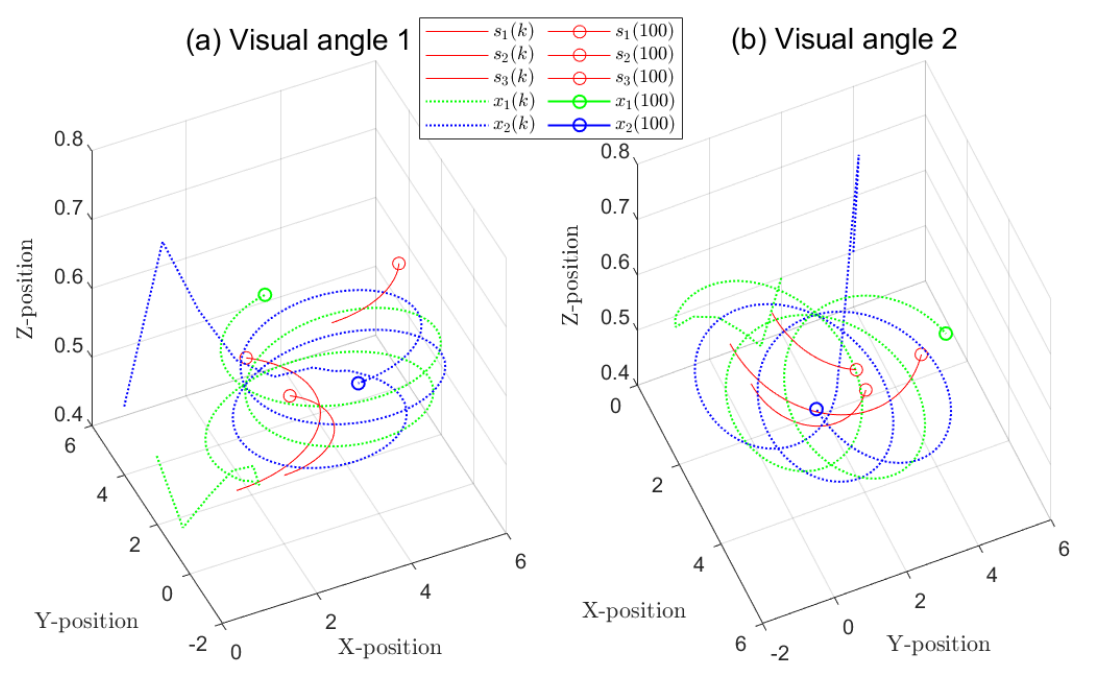}
 \caption{The simulation results in 3D system.}
  \label{3D}
\end{figure}

Besides, the initial heights of the targets are varied to further demonstrate the usefulness of the designed algorithm in the three-dimensional system, e.g., the initial position of the target 3 is reset to $\bs_{3}(0)=(3, 0, 0.7)$. Fig. \ref{3D} depicts the simulation results in this case, and we can see that the estimator and controller still work well.

\setlength{\belowcaptionskip}{-0.5cm}

\begin{figure}
\setlength{\abovecaptionskip}{-0.05cm}
\centering
  \includegraphics[width=7cm]{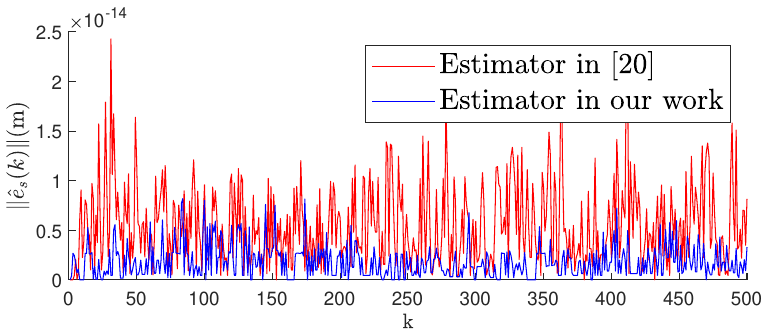}
 \caption{The estimation error for the known displacements of all targets under different estimators.}
  \label{estimator}
\end{figure}

In order to further prove the effectiveness of the proposed target-center estimator, we use the positioning algorithm in \cite{Thien2020Persistently} to compare with our algorithm under the assumption that the displacements of all targets are known. The algorithm used in \cite{Thien2020Persistently} is to estimate the position of the target by multiple steps movement of an agent, which is different from our work. In terms of the effectiveness, as shown in the Fig. \ref{estimator}, it can be seen that the error is very small under the action of both estimators (the accuracy reaches $10^{-14}$). By contrast, the estimation error is even smaller under the action of the designed estimator in this work. In particular, our work is carried out for non-cooperative unknown targets. When the displacement of the target is unknown, the estimator in \cite{Thien2020Persistently}  cannot be applied to our work.

\begin{figure}
\setlength{\abovecaptionskip}{-0.05cm}
\centering
  \includegraphics[width=7cm]{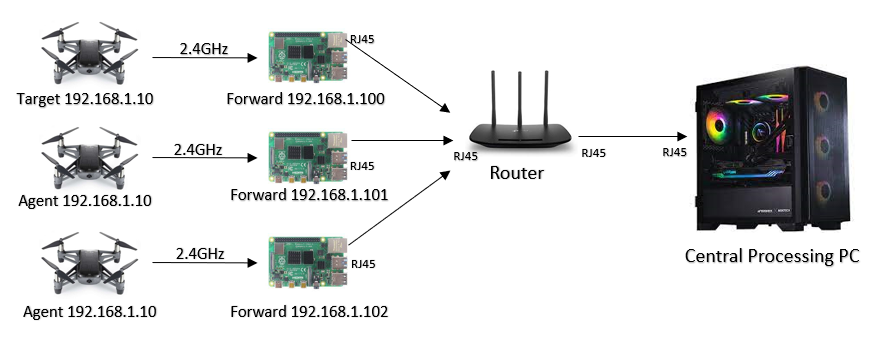}
 \caption{Overall UAV system setup.}
  \label{fig:uav_experiment_hardware_setup}
\end{figure}

\textbf{Flight experiment:}
In the physical experiment, the commercial off-the-shelf (COTS) low-cost Tello drones are deployed to represent both targets and agents. The overall hardware setup for the UAV experiment can be illustrated in Fig. \ref{fig:uav_experiment_hardware_setup}. To obtain the range information, CCMSLAM is adopted to find the relative pose relation between each ORBSLAM output pointclouds. The pointcloud scale is adjusted by Tello altitude measurement. By minimizing the reprojection error of ORBSLAM submaps from each drone, an overall global pointcloud map can be created. Since each drone is running independent scale-aware ORBSLAM, the relative position between the individual submap starting point and current pose is known. Therefore, the relative distance between each drone can be calculated. After the relative distances are obtained, the system follows the procedure as described in section \ref{sec:Problem_formulation} to form the close loop control. The overall control flow for the UAV experiment can be illustrated in Fig. \ref{fig:uav_control_feedback_loop}.

\begin{figure}
\centering
  \includegraphics[width=7cm]{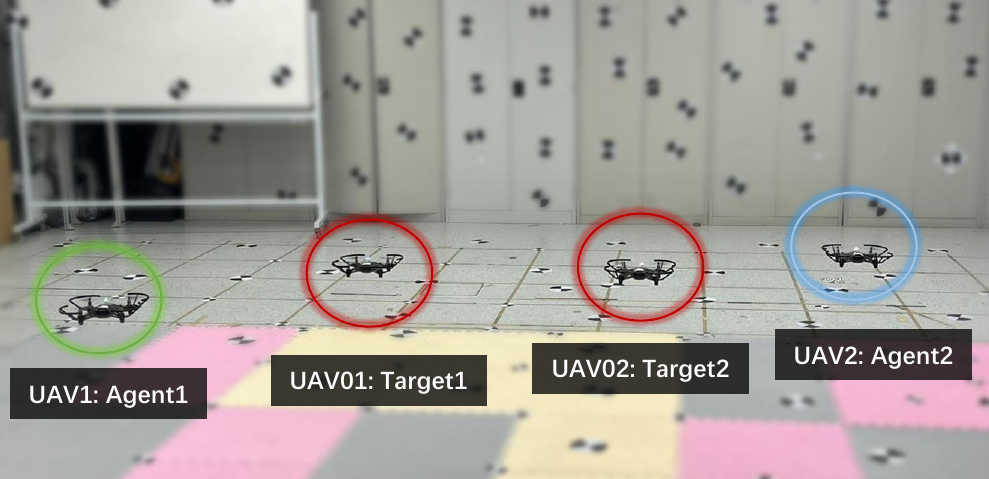}
 \caption{All UAVs for the physical experiment.}
  \label{alluavs}
\end{figure}

\begin{figure}
\setlength{\abovecaptionskip}{-0.05cm}
\centering
  \includegraphics[width=7cm]{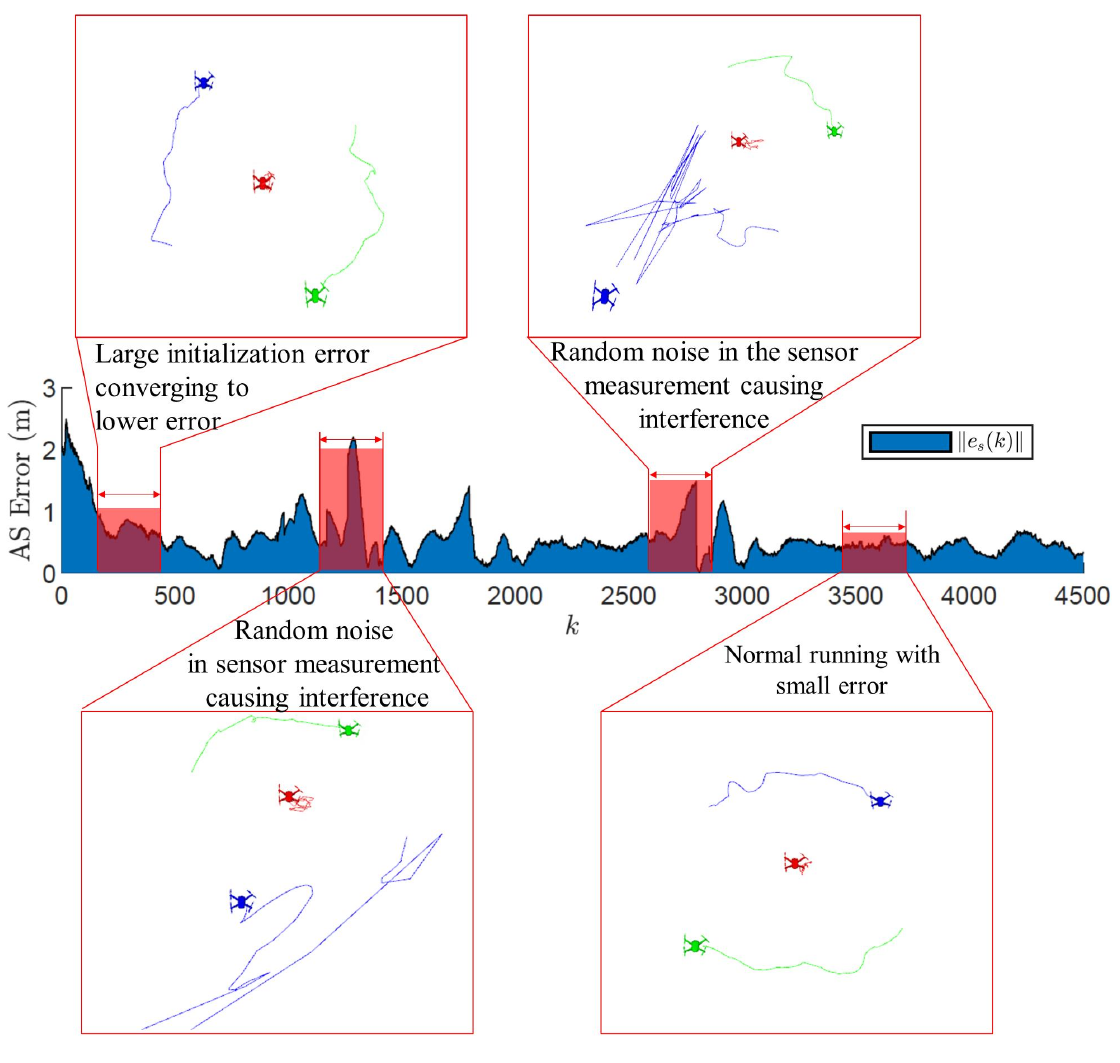}
 \caption{The AS error (blue) with respect to $k$-th frame (frame id) and the corresponding sessions of UAV flight status.}
  \label{single-error}
\end{figure}

Four UAVs are used during the experiment, with UAV01 and UAV02 identified as the targets, UAV1 as Agent 1 and UAV2 as Agent 2, which is shown in Fig. \ref{alluavs}. Both agile single-target experiments and multiple none-cooperative target experiments are conducted to demonstrate the performance of the proposed algorithm. For the single agile target study, the initial positions of all UAVs are set as (0,3), (0,2.6) and (1.7,2.3) in the $X$ axis and the $Y$ axis and fly at an altitude of 0.5 meters. Furthermore, the positions of all UAVs are sampled at around 7Hz. The experimental results for the single target are presented as shown in Fig. \ref{single-error}.  From the position trajectories of all UAVs, the UAV1 and UAV2 can track and surround UAV01 well. Moreover, the absolute value of AS error exists within a very small range in the presence of measurement error.

\begin{figure}
\setlength{\abovecaptionskip}{-0.05cm}
\centering
  \includegraphics[width=7cm]{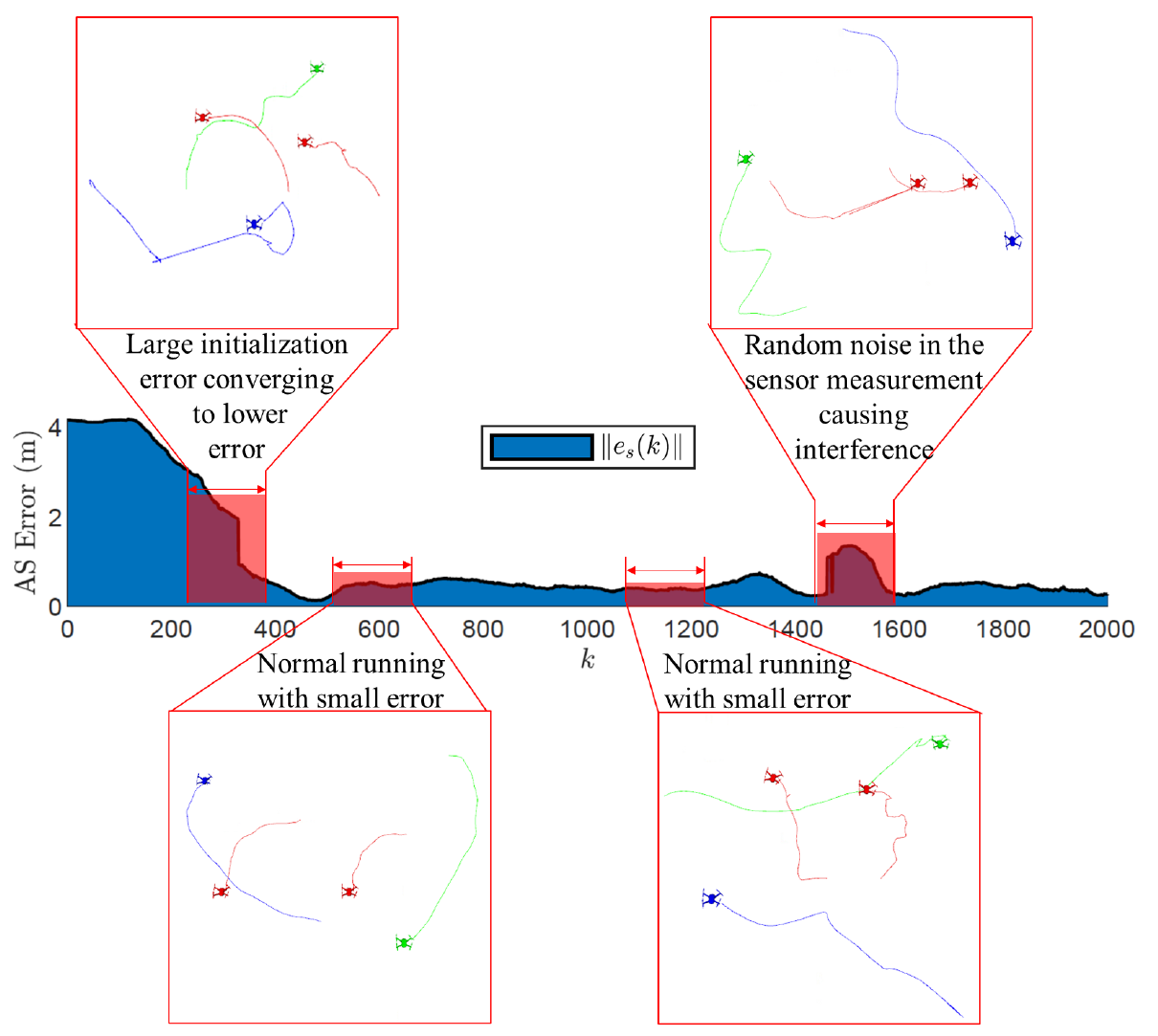}
 \caption{The AS error (blue) with respect to $k$-th frame (frame id) and the corresponding sessions of UAV flight status.}
  \label{multi-error}
\end{figure}

For the cases of multiple none-cooperative target encirclement tests, the initial positions of UAV1 and UAV2 are set as (1.2, 0) and (2.4, 0). The frequency of the preset circumnavigation trajectory is set as $\frac{1}{200}$, thus avoiding flight instability caused by the UAVs moving too fast. UAV01 and UAV02 are the targets. Under the action of the control algorithm, UAVs 1 and 2 can always be distributed on both sides of target-center (the position center of UAV01 and UAV02), and the circumnavigation of the UAV1 and UAV2 to the UAV01 and UAV02 can be realized, which can be seen in Fig. \ref{multi-error}. Moreover, it can be found from the above experimental results that the absolute value of the AS error is less than 0.8 in most of the time, which means that the target-centered AS is better achieved, so that the target (UAV0) is covered by the sensors of the other two UAVs. Detailed URL link for video demonstration can be found in our abstract section.

\setlength{\belowcaptionskip}{-0.5cm}

\section{conclusion}
In this work, we have studied the multiple non-cooperative targets encirclement problem. Based only on the measured distances and relying on tools such as FWNN and the least squares fit, the center of all targets has been directly estimated. In order to cover all targets to the maximum extent, the radius of the enclosing trajectory has been intelligently adjusted according to the distance between each target. Sufficient conditions for the estimator and controller to work have been obtained. Furthermore, the effectiveness of the target-center position estimator and DASC has been further demonstrated by a numerical simulation and a UAV-based experimental evaluation. The case of targets with stochastic dynamics can be considered in future work.

\vspace{-0.1cm}

\bibliographystyle{Bibliography/IEEEtranTIE}
\bibliography{reference}

\begin{IEEEbiography}
[{\includegraphics[width=1in,height=1.25in]{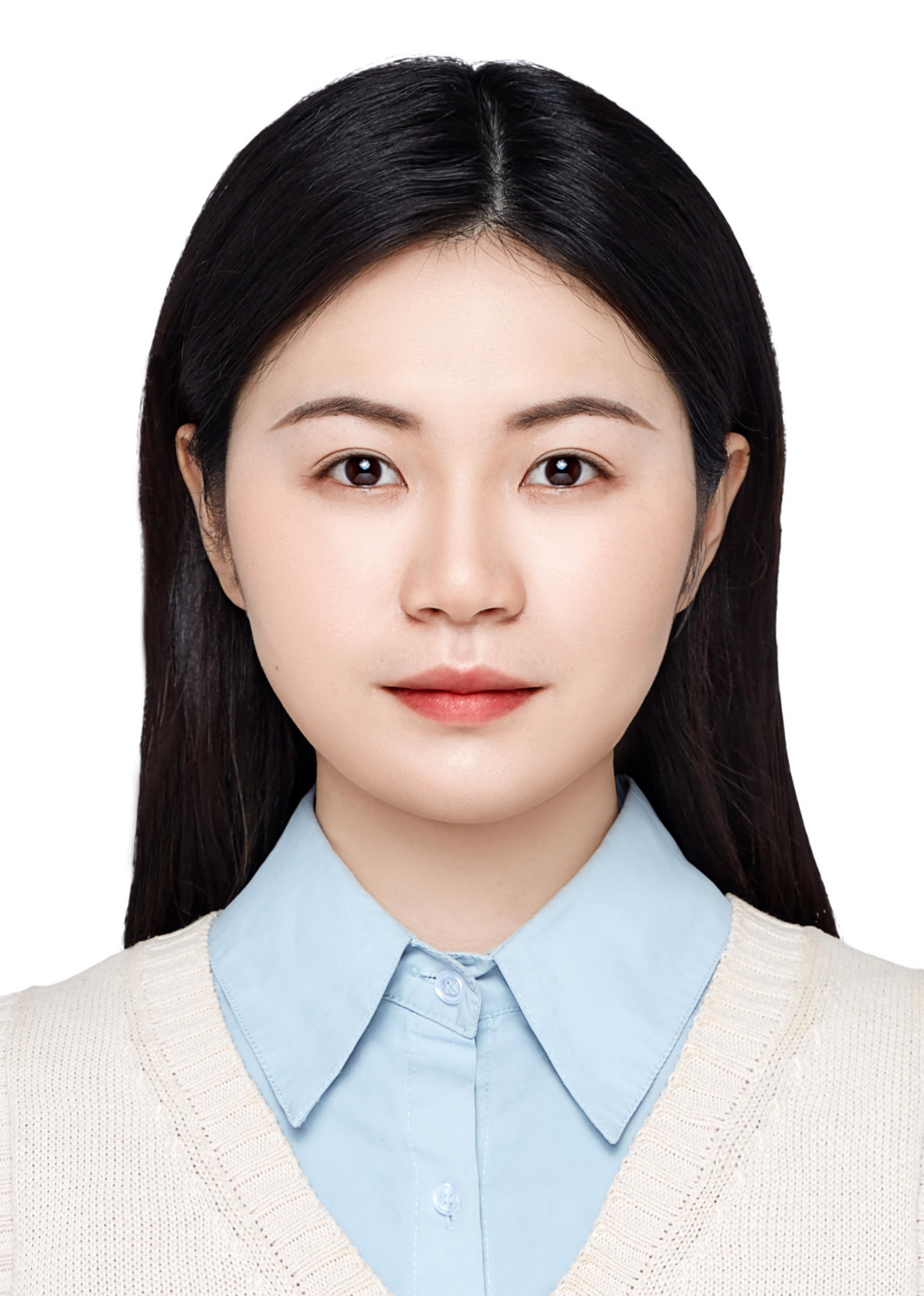}}]{Fen Liu} received the M.S. degree in 2020, and is currently pursuing the Ph.D. degree, both in the School of Automation, Guangdong University of Technology, Guangzhou, China. Meanwhile, she is currently a visiting student at the School of Electrical and Electronic Engineering, Nanyang Technological University, Singapore. Her current research interests include the target encirclement, cooperative control, anti-synchronization control.
\end{IEEEbiography}

\begin{IEEEbiography}
[{\includegraphics[width=1in,height=1.25in]{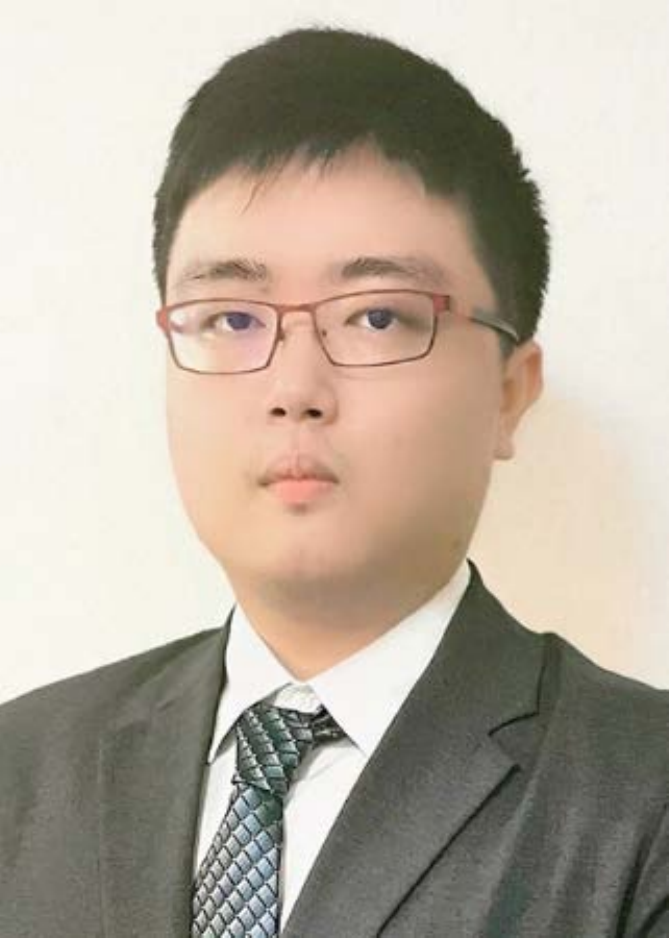}}]{Shenghai Yuan} received his Bachelor and Ph.D. degrees in electrical and electronic engineering from The Nanyang Technological University, Singapore, in 2019 and 2013, respectively. He is a Research Fellow in Center for Advanced Robotics Technology Innovation (CARTIN), Nanyang Technological University. His research interests include the areas of perception, Sensor fusion for robust navigation, machine learning and various autonomous system. \\
\end{IEEEbiography}

\begin{IEEEbiography}
[{\includegraphics[width=1in,height=1.25in]{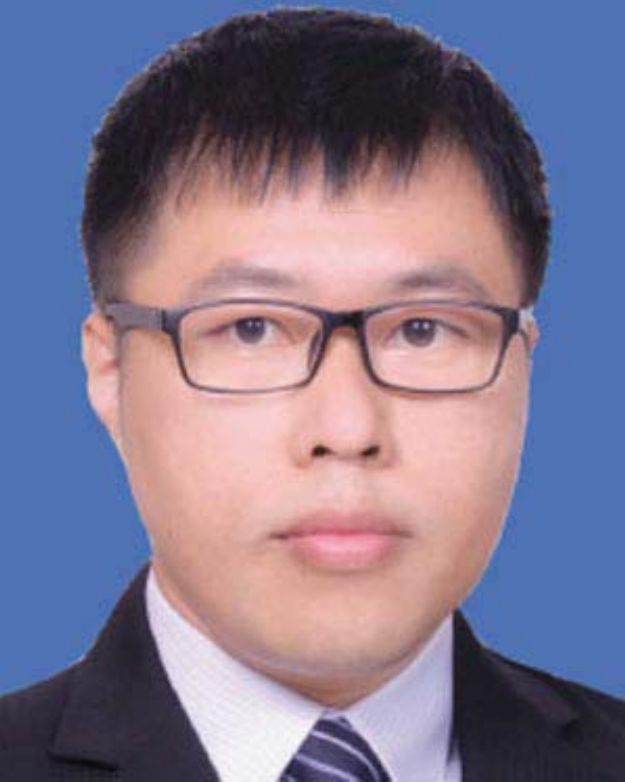}}]{Wei Meng} received the B.E.
and M.E. degrees from Northeastern University, Shenyang, China, in 2006 and 2008, respectively, both in electrical engineering, and the Ph.D. degree in control and instrumentation from the Nanyang Technological University, Singapore,
in 2013. From 2012 to 2017, he was a Research Scientist with UAV Research Group, Temasek Laboratories, National University of Singapore. He is currently with the School of Automation, Guangdong University of Technology as a Professor. His current research interests include unmanned systems, cooperative control, multirobot systems, localization, and tracking.
\end{IEEEbiography}

\begin{IEEEbiography}
[{\includegraphics[width=1in,height=1.25in]{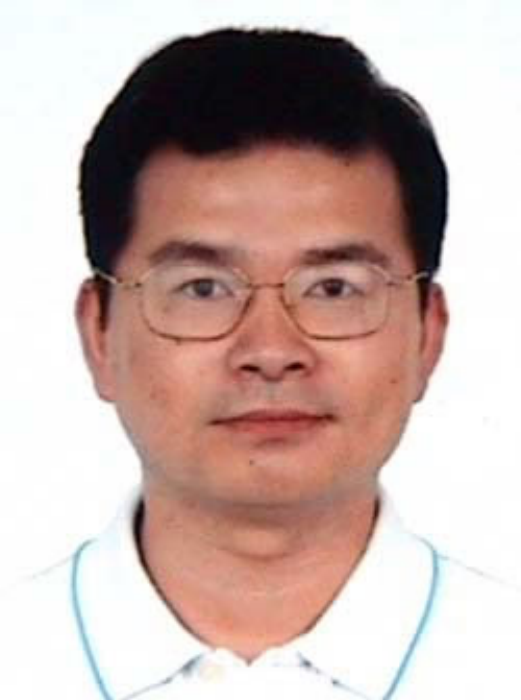}}]{Rong Su} obtained his Bachelor of Engineering degree from University of Science and Technology of China in 1997, and Master of Applied Science degree and PhD degree from University of Toronto, in 2000 and 2004, respectively. He was affiliated with University of Waterloo and Technical University of Eindhoven before he joined Nanyang Technological University in 2010. Currently, he is an associate professor in the School of Electrical and Electronic Engineering. His research interests include multi-agent systems, discrete-event system theory, model-based fault diagnosis, operation planning and scheduling with applications in flexible manufacturing, intelligent transportation, human-robot interface, power management and green building.\\
\end{IEEEbiography}

\begin{IEEEbiography}
[{\includegraphics[width=1in,height=1.25in]{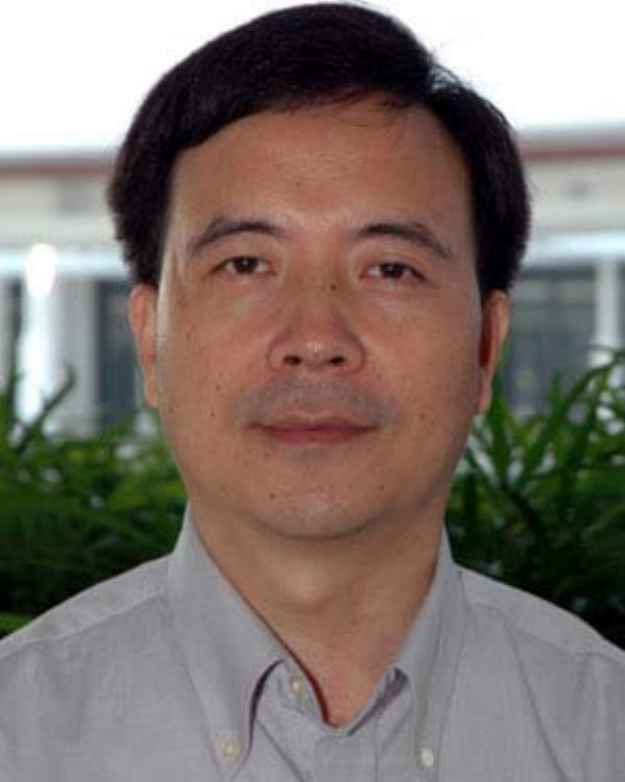}}]{Lihua Xie} (Fellow, IEEE) received his Ph.D. degree in electrical engineering from The University of Newcastle, Australia, in 1992 and B.E. and M.E. degrees in control engineering from Nanjing University of Science and Technology in 1983 and 1986, respectively. He joined The Nanyang Technological University, Singapore, in 1992 where he is currently a Professor at the School of Electrical and Electronic Engineering and Director at Delta-NTU Corporate Laboratory for Cyber-Physical Systems. His current research interests include robust control, networked control, multi-agent systems, sensor networks, energy efficient buildings, and unmanned systems. In these areas, he has published over 800 technical papers including over 440 journal papers and coauthored five patents and nine books. Lihua Xie is Fellow of Academy of Engineering Singapore, Fellow of IEEE, and Fellow of IFAC. \\
\end{IEEEbiography}
\end{document}